\documentclass[letterpaper,10pt,conference]{ieeeconf}

\IEEEoverridecommandlockouts
\usepackage{amsmath}
\usepackage{amssymb}
\usepackage{booktabs}
\usepackage{cite}
\usepackage{float}
\usepackage{graphicx}
\usepackage{multirow}
\usepackage{xcolor}
\usepackage[hidelinks]{hyperref}
\usepackage{url}

\makeatletter
\renewcommand{\subsection}{\@startsection{subsection}{2}{\z@}%
  {1.5ex plus 1.5ex minus 0.5ex}%
  {0.9ex plus 0.5ex minus 0ex}%
  {\normalfont\normalsize\itshape}}
\makeatother

\definecolor{teacherblue}{RGB}{42,111,151}
\definecolor{studentorange}{RGB}{219,120,32}
\definecolor{softgray}{RGB}{241,243,245}
\definecolor{lossred}{RGB}{174,55,55}

\newcommand{\piZero}{\ensuremath{\pi_{0}}}
\newcommand{\piHalf}{\ensuremath{\pi_{0.5}}}
\newcommand{\method}{ForeTime-VLA}
\newcommand{\best}[1]{\textbf{#1}}

\title{\LARGE \bf
ForeTime-VLA: Causal Future-Token Distillation from a
World Action Model for Conveyor-Belt Manipulation
}

\author{
  \authorblockN{Siyuan Ma\textsuperscript{*,1}, Yutian Zhang\textsuperscript{*,2},
  Boshi Zhang\textsuperscript{*,1}, Qinglian Wu\textsuperscript{3},\\
  Jiaqi Zhai\textsuperscript{4}, Dong Wei\textsuperscript{\textdagger,4},
  Xiaojin Huang\textsuperscript{\textdagger,1}}
  \authorblockA{\textsuperscript{1}Tsinghua University, Beijing, China\\
  \textsuperscript{2}Shanghai Artificial Intelligence Laboratory, Shanghai, China\\
  \textsuperscript{3}Harbin Institute of Technology, Harbin, China\\
  \textsuperscript{4}Hangzhou Yunshenchu Technology Co., Ltd. (DEEP Robotics), Hangzhou, China\\
  \textsuperscript{*}Equal contribution. \textsuperscript{\textdagger}Corresponding authors.}
}

\begin{document}

\bstctlcite{BSTcontrol}
\maketitle
\thispagestyle{empty}
\pagestyle{empty}

\begin{abstract}
Manipulating moving objects requires a policy to anticipate contact events, yet
vision-language-action (VLA) policies are commonly fine-tuned from the current
observation alone.  World action models (WAMs) learn predictive dynamics, but
running a video-scale teacher or explicitly imagining future frames at deployment
is costly.  We introduce \method, a dense \piHalf\ policy that distills a
future-aware, action-equivalent representation from a frozen Fast-WAM-derived
teacher while remaining causal at inference.  Offline, current and future video
latents are compressed into a whitened 64-D target.  Online, an eight-frame
history encoder predicts this target together with manipulation phase and
normalized time-to-transition.  Four future tokens and one phase token condition
the VLM prefix, while the predicted future and transition horizon condition the
action expert.  Training
retains the original flow-matching action target and adds cosine, relational
geometry, phase, time-to-transition, and action-equivalence objectives.  On a deduplicated
conveyor-belt dataset, we compare models on 768 matched windows per split.  Test
MAE decreases from 0.134119 to 0.130593 (2.63\%; paired-bootstrap
95\% CI: 0.82--4.48\% improvement), and test L2 decreases by 3.02\%, at a
2.46--2.93\% latency cost.  In quantitative real-robot evaluation, \method
achieves 81.1\% stationary and 58.9\% slow-moving grasp success, exceeding the
next-best reference by 12.2 and 22.2 percentage points, respectively.  Across
three belt speeds, it completes 44/90 grasps versus 23/90 for \piHalf,
including 11/30 versus 2/30 at fast speed.  The agreement between offline
orientation gains and reduced real-robot contact-pose failures supports causal
future-token distillation as an effective way to improve dynamic manipulation
without deploying the world-model teacher.
\end{abstract}

\section{Introduction}
\label{sec:intro}

Vision-language-action (VLA) models transfer semantic priors from large
vision-language backbones to robot control.  RT-2~\cite{zitkovich2023rt2},
OpenVLA~\cite{kim2024openvla}, Octo~\cite{octoteam2024octo}, and the
\piZero/\piHalf\ family~\cite{black2024pi0,physicalintelligence2025pi05} show
that a common policy can be adapted across tasks, scenes, and embodiments.
Expressive action decoders---including diffusion~\cite{chi2023diffusionpolicy}
and flow matching~\cite{lipman2023flowmatching}---further support multimodal,
high-dimensional action chunks.

Dynamic conveyor-belt pick-and-place exposes a complementary weakness.  The
correct motion depends not only on what is visible now, but also on when the
object will enter the reachable set, when the gripper should close, and how the
end effector should be oriented at contact.  A policy trained only to reconstruct
demonstrated actions may learn these regularities indirectly, but receives no
explicit future-structure supervision.

World models provide such supervision through video prediction.  Text-guided
video policies~\cite{du2023unipi}, generative video-language-action
models~\cite{wu2023gr1}, and recent world action models (WAMs) use predicted
future observations to support control.  Explicit imagine-then-act pipelines,
however, add video generation to the control loop.  Fast-WAM~\cite{yuan2026fastwam}
instead reports that video modeling is especially useful as a training signal
and can be removed from test-time inference.  This raises a practical question:
\emph{can a pretrained VLA acquire a compact version of that predictive
structure without running the WAM at deployment?}

\begin{figure*}[t]
\centering
\includegraphics[width=0.98\textwidth]{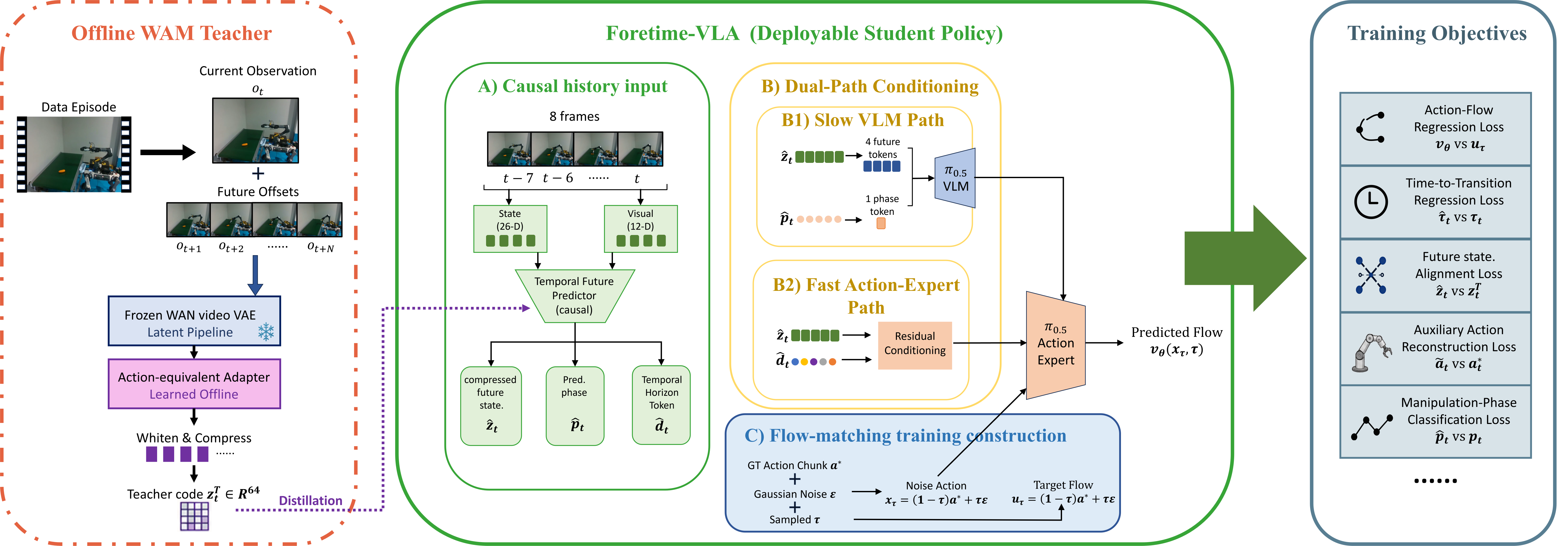}
\caption{\textbf{\method\ overview.} \emph{Left:} the offline teacher encodes
the current observation and sampled future frames with a frozen video VAE, then
maps them to a compact action-equivalent target. \emph{Center:} the deployable
student uses eight causal observations to predict future, phase, and transition
horizon variables. These predictions condition both the slow \piHalf\ VLM
prefix and the fast action-expert suffix, while the original noisy-action and
target-flow construction is retained. \emph{Right:} training combines action
flow, future alignment, relational geometry, transition, phase, and auxiliary
action-reconstruction losses. Future frames and the WAM are absent at test
time.}
\label{fig:method}
\end{figure*}

We answer this question with \method, whose teacher--student pipeline is
summarized in Fig.~\ref{fig:method}.  During offline preprocessing, a
Fast-WAM/Wan video latent pipeline sees the current observation and eight
future offsets.  A non-collapsed adapter compresses these features into an
action-equivalent teacher code.  During policy training, a small causal encoder
must predict that code from the preceding eight observations.  The predicted
code conditions both the slow VLM path and the fast action-expert path of
\piHalf, while the original flow-matching construction continues to use the
recorded action chunk.  Ground-truth actions remain unchanged, avoiding the
target-alignment confound of pseudo-label replacement.

Our contributions are:
\begin{itemize}
    \item a deployable teacher--student formulation that transfers
    future-conditioned WAM structure into an eight-frame causal history encoder,
    with no future frame or teacher forward pass at inference;
    \item dual-path conditioning of \piHalf\ with future, phase, and
    time-to-transition variables, supervised by pointwise, relational, temporal, and
    action-equivalence objectives; and
    \item a paired offline evaluation and two quantitative real-robot studies
    spanning stationary, moving, and speed-stress conditions, with outcome-level
    analysis connecting future-aware supervision to fewer late and contact-pose
    failures.
\end{itemize}

\section{Related Work}
\label{sec:related}

\subsection{Vision-Language-Action Policies}

VLA policies connect visual-language representations to robot actions.
RT-1 demonstrated large-scale transformer control from image and language
inputs~\cite{brohan2022rt1}, while Gato and PaLM-E placed robot control within
broader generalist and embodied multimodal models~\cite{reed2022gato,driess2023palme}.
The Open X-Embodiment effort subsequently assembled cross-embodiment data and
RT-X policies to study transfer across robot platforms~\cite{padalkar2023openx}.
RT-2 casts actions as language tokens~\cite{zitkovich2023rt2}; OpenVLA provides
an open 7B policy and efficient adaptation recipe~\cite{kim2024openvla}; and
Octo studies flexible observation and action interfaces~\cite{octoteam2024octo}.
\piZero\ introduces a VLM-conditioned action expert trained with flow
matching~\cite{black2024pi0}.  \piHalf\ extends this family with heterogeneous
co-training and high-level semantic prediction~\cite{physicalintelligence2025pi05}.
GR00T N1 couples a vision-language backbone to a diffusion-transformer action
head and provides an open recipe for adapting the policy to new
embodiments~\cite{bjorck2025gr00t}.
FAST explores efficient action tokenization for high-frequency
control~\cite{pertsch2025fast}.  Complementary manipulation systems study
multimodal prompting, multi-task 3-D action prediction, and long-horizon
language-conditioned evaluation~\cite{jiang2023vima,shridhar2023peract,mees2022calvin}.
Our work keeps the continuous flow-matching
action interface of \piHalf\ and adds predictive temporal supervision during
full-parameter fine-tuning.

\subsection{Predictive Representations for Control}

Latent world models learn compact predictive state and support behavior through
imagined rollouts~\cite{ha2018worldmodels,hafner2020dreamer,hafner2023dreamerv3}.
In robotics, visual-foresight methods predict future observations for model-based
planning~\cite{finn2017deepvisual,ebert2018visualforesight}, while UniSim and
Genie explore generative interactive environments at broader visual
scales~\cite{yang2023unisim,bruce2024genie}.
Video-conditioned policies can use generated futures as plans or predictive
features~\cite{du2023unipi,wu2023gr1,hu2024vpp}.  Fast-WAM separates video
co-training from future synthesis and argues that predictive representation
learning can retain much of the WAM benefit without test-time
imagination~\cite{yuan2026fastwam}.  DECOWAM extends this direction to legged
mobile manipulation by decoupling camera ego-motion, base actions, and arm
actions while distilling an action-equivalent future bottleneck from privileged
observations~\cite{ma2026decowam}.  \method\ is complementary: rather than
deploying the video-scale WAM, it distills a 64-D future-aware code into a
pretrained VLA and predicts the code from compact causal histories.

\subsection{Knowledge Distillation}

Classical knowledge distillation transfers the behavior of a costly teacher to
a deployable student~\cite{hinton2015distilling}.  Intermediate-feature and
relational distillation preserve more than output logits; relational knowledge
distillation, for example, matches structure between examples in representation
space~\cite{romero2015fitnets,zagoruyko2017attention,park2019rkd,tian2020crd}.
Because intermediate targets can become low-rank or collapsed, variance and
covariance regularization provide useful complementary constraints
~\cite{bardes2022vicreg,jing2022collapse}.  Our future cosine loss transfers
instance-level direction, while a Gram-matrix loss transfers batch geometry.
Phase, time-to-transition,
and action reconstruction constrain the code to remain useful for manipulation
rather than merely matching an arbitrary latent basis.

\section{Method}
\label{sec:method}

\subsection{Problem Formulation}

At time $t$, the policy receives current RGB observations $o_t$, a language
instruction $\ell$, robot state $s_t$, and an eight-step causal history
$\mathcal{H}_t$.  It predicts an action chunk
$a_{t:t+H-1}\in\mathbb{R}^{H\times D}$, with $H=20$ and $D=16$.  The base
\piHalf\ action expert uses conditional flow matching.  For actions $a$, noise
$\epsilon\sim\mathcal{N}(0,I)$, and $\tau\in(0,1)$,
\begin{equation}
    x_\tau=(1-\tau)a+\tau\epsilon,\qquad u_\tau=\epsilon-a ,
\end{equation}
and the base loss is
\begin{equation}
    \mathcal{L}_{\mathrm{flow}}
    =\mathbb{E}\left[\left\|v_\theta(x_\tau,\tau,o_t,\ell)-u_\tau\right\|_2^2\right].
\end{equation}
This noisy-action/target-flow construction is shown in
Fig.~\ref{fig:method}(C) and is preserved when the temporal conditioning is added.
Our goal is to add future-sensitive structure without changing $a$ and without
using future observations at inference.

\subsection{Offline Action-Equivalent Future Teacher}

The left side of Fig.~\ref{fig:method} depicts the privileged teacher branch.
For every valid training window, we sample the present and eight future frames
at offsets $\{2,4,7,9,12,14,17,19\}$.  A frozen Wan2.2 video
VAE~\cite{wan2025wan}---the visual latent pipeline used by
Fast-WAM~\cite{yuan2026fastwam}---encodes the sequence.  Spatial-temporal means
and standard deviations produce 96-D current and 96-D future features.

The initially exported quotient codes exhibited insufficient variation.
We therefore train a non-collapsed action-equivalent adapter while keeping the
video features frozen.  Its teacher encoder consumes current feature, future
feature, and the 26-D state; a causal auxiliary encoder consumes only current
feature and state.  Both emit 64-D codes.  A shared action decoder reconstructs
the normalized $20\times16$ action chunk, and a future decoder reconstructs the
future-minus-current feature.  Training combines teacher and causal action MSE,
future smooth-L1, teacher--causal cosine, per-dimension variance, off-diagonal
covariance, and pairwise action-geometry losses with weights
$1.0$, $0.5$, $0.25$, $0.10$, $0.20$, $0.01$, and $0.05$, respectively.
The resulting teacher codes are whitened per dimension and cached.
Thus the privileged future is used to define $z_t^T$, but never enters the
deployed policy.

\begin{figure}[t]
\centering
\includegraphics[width=0.98\columnwidth]{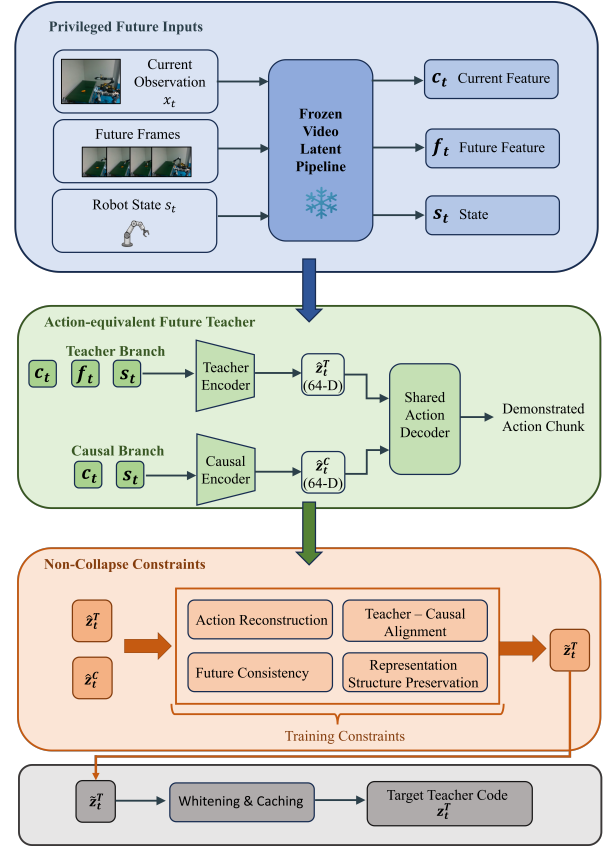}
\caption{\textbf{Offline action-equivalent future teacher.} The frozen latent
pipeline extracts current and future features $(c_t,f_t)$, and the adapter also
receives robot state $s_t$. The privileged encoder uses $(c_t,f_t,s_t)$,
whereas the causal branch uses $(c_t,s_t)$. Both codes share an action decoder;
the teacher additionally reconstructs the future feature difference.
Teacher--causal alignment, variance/covariance regularization, and pairwise
action geometry prevent collapse and retain action relevance. The selected
64-D teacher code is whitened and cached as $z_t^T$.}
\label{fig:teacher_adapter}
\end{figure}

Figure~\ref{fig:teacher_adapter} summarizes the adapter from top to bottom.  The
frozen latent pipeline extracts $c_t$ from the current observation and $f_t$ from
the sampled future frames, while the 26-D state $s_t$ is supplied to the
adapter.  The privileged teacher encoder uses all three inputs,
$\hat z_t^T=g_T(c_t,f_t,s_t)$, whereas the causal auxiliary encoder uses only
$(c_t,s_t)$ and produces $\hat z_t^C=g_C(c_t,s_t)$.  Both 64-D codes are sent
through a shared action decoder, and the teacher code is additionally sent to a
future decoder that reconstructs $f_t-c_t$.  Teacher--causal alignment transfers
future information to the causal branch; variance and covariance regularization
keep every latent dimension active; and pairwise action geometry preserves
relative distances between examples.  After training, only the per-dimension
whitened teacher output $z_t^T$ is cached for causal-student supervision.

\subsection{Causal Temporal Student}

Figure~\ref{fig:method}(A) shows the causal input and its three prediction heads.
The deployment history contains eight 26-D normalized states and 12-D visual
statistics (per-channel means and standard deviations from two auxiliary
cameras).  The current full-resolution images still enter the standard
\piHalf\ vision backbone.  For history index $i$, we compute
\begin{equation}
 e_i=\operatorname{swish}(W_s\operatorname{LN}(s_i))
    +\operatorname{swish}(W_v\operatorname{LN}(r_i))+p_i ,
\end{equation}
where $p_i$ is a learned temporal embedding and $e_i\in\mathbb{R}^{256}$.
After flattening, a two-layer residual MLP produces
\begin{equation}
 q=\operatorname{swish}(W_p\operatorname{vec}(e_{1:8})),\quad
 h_t=\operatorname{LN}(\operatorname{swish}(W_o q)+q).
\end{equation}
Linear heads predict the compressed future state
$\hat z_t\in\mathbb{R}^{64}$, four manipulation-phase logits $\hat p_t$, and a
sigmoid normalized time-to-transition $\hat d_t\in[0,1]$; the latter is the
temporal-horizon signal shown in the figure.

The phase target is obtained from gripper-state transitions: approach,
grasp/release transition, transport while holding, and place/retract.  The
time-to-transition target is the clipped, horizon-normalized distance to the
next gripper-state transition.
These low-cost labels expose event structure that is otherwise implicit in the
action chunk.

\subsection{Dual-Path Conditioning}

As illustrated in Fig.~\ref{fig:method}(B), the student conditions the policy
through two complementary routes.  The slow path maps $\hat z_t$ to four
2048-D future tokens.  The expected
embedding under the predicted phase distribution supplies a fifth token.  The
five tokens are appended to the \piHalf\ prefix.  The fast path maps
$\hat z_t$ and $\hat d_t$ to a 1024-D residual that is added to every
action-expert suffix token and its adaptive normalization condition.  Learned
scalar gates initialize both injections at $0.05$, limiting disruption of the
pretrained policy early in training.

The added modules contain 1.251M parameters, small relative to the dense
Gemma-2B VLM and 311M-parameter action expert.  The Fast-WAM teacher, video
VAE, and offline adapter are absent from deployment.

\subsection{Training Objective}

The right side of Fig.~\ref{fig:method} groups the supervision into five
functional objectives: action flow, time to transition, future-state alignment,
auxiliary action reconstruction, and manipulation-phase classification.  The
future-state term contains both pointwise and relational components.  Let
row-normalized predicted and teacher codes in a batch be $\bar Z$ and
$\bar Z^T$.  We use pointwise cosine and relational geometry losses,
\begin{align}
 \mathcal{L}_{\mathrm{cos}}
    &=\frac{1}{B}\sum_i(1-\bar z_i^\top\bar z_i^T),\\
 \mathcal{L}_{\mathrm{geo}}
    &=\frac{1}{B^2}\left\|\bar Z\bar Z^\top
                         -\bar Z^T(\bar Z^T)^\top\right\|_F^2 .
\end{align}
Cross-entropy supervises phase, Huber loss with $\delta=0.1$ supervises
time to transition, and a linear decoder from $\hat z_t$ reconstructs the normalized
action chunk for $\mathcal{L}_{\mathrm{ae}}$.  The total objective is
\begin{equation}
\begin{split}
\mathcal{L}={}&\mathcal{L}_{\mathrm{flow}}
+0.20\mathcal{L}_{\mathrm{cos}}
+0.02\mathcal{L}_{\mathrm{geo}}
+0.04\mathcal{L}_{\mathrm{phase}}\\
&+0.04\mathcal{L}_{\mathrm{transition}}
+0.05\mathcal{L}_{\mathrm{ae}} .
\end{split}
\end{equation}
All \piHalf\ parameters and \method\ modules are optimized jointly; the
recorded action target is never replaced by a teacher action.

\subsection{Design Rationale and Distinction}

\method\ transfers predictive structure rather than generated pixels.  Its
privileged branch compresses future video into a code constrained by the
demonstrated action, filtering appearance changes that are irrelevant to
control.  The causal student recovers this code from history alone, retaining
the temporal bias of a video model without future synthesis in the control loop.
Future and phase tokens expose the forecast to the semantic VLM path, while the
fast residual gives the action expert direct access to the predicted code and
transition horizon.  This action-equivalent, event-aware interface---together
with an unchanged action target---distinguishes \method\ from attaching a video
generator to the policy and makes it complementary to flow matching.

\section{Experimental Setup}
\label{sec:setup}

\subsection{Dataset}

\begin{figure*}[t]
\centering
\includegraphics[width=0.98\textwidth]{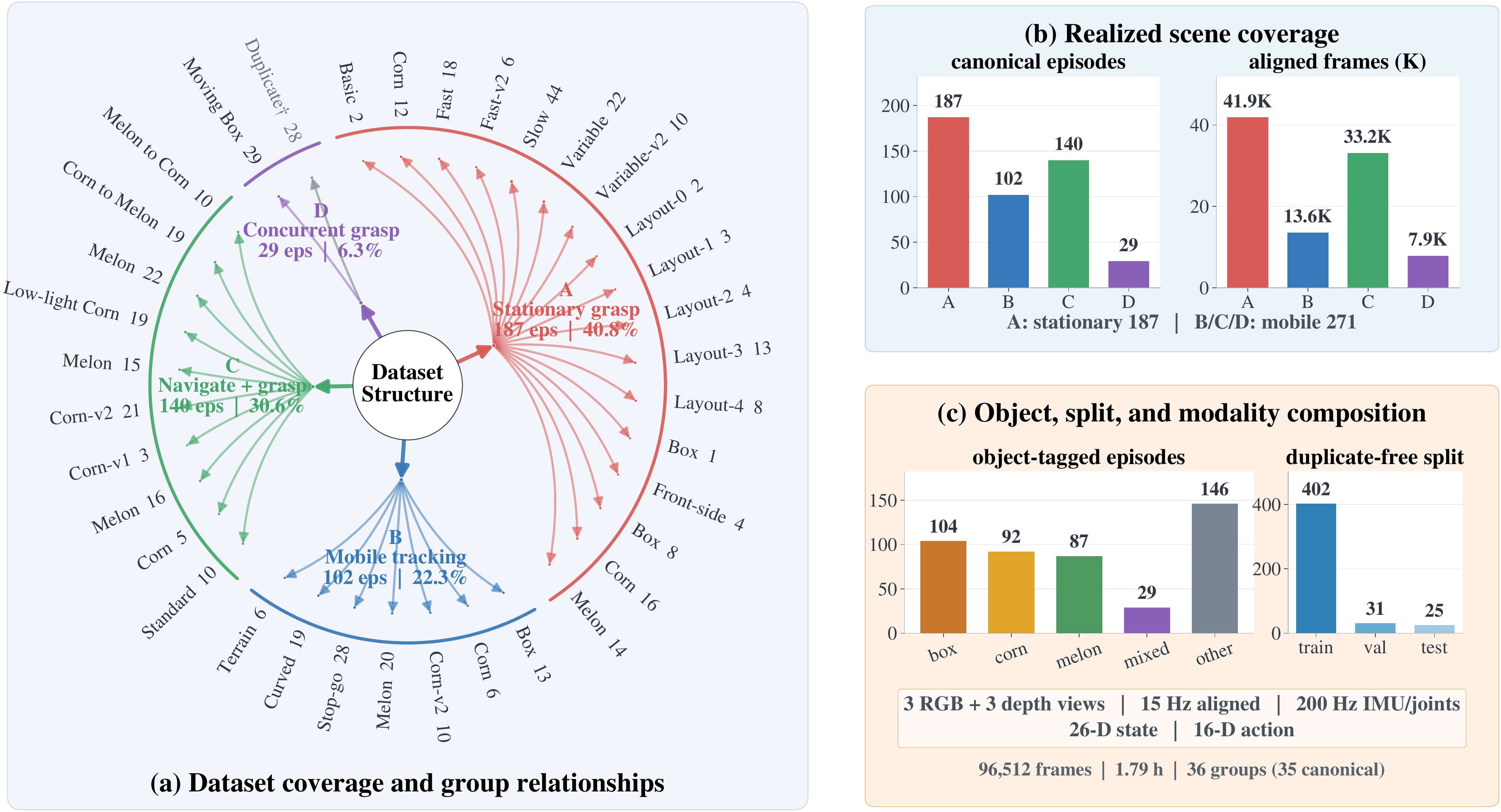}
\caption{\textbf{Dataset structure and realized coverage.} \emph{Left:} the
collection taxonomy spans stationary grasp, mobile tracking,
navigate-then-grasp, and concurrent-grasp settings; outer labels give canonical
episode counts. \emph{Upper right:} episode and aligned-frame coverage by scene.
\emph{Lower right:} object composition, the duplicate-free split, and the
recorded multimodal contract. Aggregate statistics exclude 28 exact duplicate
files; the remaining 458 episodes contain 96,512 aligned frames and are split
402/31/25 for train/validation/test.}
\label{fig:dataset}
\end{figure*}

The conveyor-belt corpus contains mobile-manipulator demonstrations of picking
boxes, corn, and watermelons under variations in belt speed, viewpoint, and
scene layout.  Auditing found 487 HDF5 files, of which one was unreadable and
28 were exact duplicates.  The remaining 458 episodes contain 96,512 frames
(1.79 hours at 15 Hz).  We use a deterministic, duplicate-free split of
402/31/25 episodes for train/validation/test, yielding 77,125/6,075/4,610
valid action windows.  All blank instructions are replaced with
``Pick up and place the moving object from the conveyor belt.''
Figure~\ref{fig:dataset} summarizes the realized collection rather than the
larger acquisition target: the outer taxonomy retains every on-disk group,
while all aggregate counts use only canonical episodes.

\subsection{Models and Protocol}

The baseline and \method\ use the same \piHalf\ backbone and base initialization.
The baseline retains the original current-observation interface, whereas
\method\ adds the temporal module and cached auxiliary supervision described in
Sec.~\ref{sec:method}.

The cross-family comparison includes GR00T N1.6-3B~\cite{bjorck2025gr00t},
StarVLA QwenGR00T, and SmolVLA-450M~\cite{shukor2025smolvla}.  All references use
the same duplicate-free data split and preserve the three camera streams, 26-D
state, 16-D absolute action, 20-step horizon, and instruction, providing a
consistent data and action contract across model families.

The shared backbone, initialization, action targets, and matched windows make
the direct \piHalf\ comparison a controlled test of temporal conditioning.  The
additional families are not parameter-matched ablations, but test whether the
benefit remains competitive beyond a single baseline family.

For each split, the evaluator draws 768 windows.  All five models receive
exactly matched episode/start indices and raw target action chunks.  We report
mean squared error (MSE), mean absolute error (MAE), and mean per-timestep
action-vector L2 error in the unnormalized 16-D action space; lower is better.
Confidence intervals for the controlled \piHalf\ comparison use 20,000 paired
bootstrap resamples; relative improvement is
$100(m_{\mathrm{base}}-m_{\mathrm{ours}})/m_{\mathrm{base}}$.
Latency is measured on the same A100.

\subsection{Real-Robot Protocol}

The reported spreadsheet contains two closed-loop real-robot evaluations.  The
first compares all five policies on a stationary task and a slow-speed moving
task, with 90 trials per method in each task.  The stationary task reports grasp
success.  For the moving task, every trial is assigned to exactly one of four
outcomes: grasp success, early failure, late failure, or contact-pose error; the
four counts therefore sum to 90 for each method.

The second evaluation isolates sensitivity to belt speed.  It compares the
\piHalf\ baseline and \method\ over slow, medium, and fast settings, with 30
trials per method at each speed.  We report the exact workbook counts and their
corresponding rates.

\section{Results}
\label{sec:results}

\subsection{Aggregate Reconstruction}

Table~\ref{tab:main} reports validation and test MAE/L2, the metrics for which
the paired evidence is strongest.  Validation and test MAE improve by 2.96\%
and 2.63\%, respectively, with 95\% intervals above zero, while test L2 improves
by 3.02\%.  The consistent validation/test gains show that the distilled future
representation improves action reconstruction beyond the training distribution,
with especially clear benefits for coordinate-wise and action-vector accuracy.

The confidence intervals bound this claim: evidence is positive for MAE on both
splits and for test L2, whereas validation L2 and the separately analyzed MSE
intervals overlap zero.  More importantly,
the offline gains agree with independent downstream evidence: orientation
improves most, component masking harms transition accuracy, and the closed-loop
advantage grows as motion shortens the interception horizon.  Agreement across
reconstruction, intervention, and execution is stronger evidence for useful
temporal foresight than any single aggregate score.

MSE is retained as a complementary view of error concentration.  ForeTime-VLA
has the best validation MSE (0.249276), while StarVLA has the lowest test MSE
(0.237490); the latter does not overturn the broader result because ForeTime-VLA
remains best on both test MAE and test L2.  In other words, the proposed policy
reduces typical coordinate and action-vector error consistently across splits,
whereas the MSE ranking is more sensitive to a small number of large residuals.
The paired MSE intervals also cross zero on validation and test, so we do not
claim a statistically significant MSE advantage.  Reporting all three metrics
therefore makes the comparison more informative: MSE exposes tail sensitivity,
while MAE and L2 capture the stable accuracy gains that align with the robot
results.

\begin{table}[H]
\caption{Offline MSE, MAE, and L2 over 768 matched windows per split. Lower is
better. The lower panel reports paired relative improvement and 95\% CIs for
\method\ versus the \piHalf\ baseline.}
\label{tab:main}
\centering
\scriptsize
\setlength{\tabcolsep}{3pt}
\begin{tabular}{@{}llrrr@{}}
\toprule
Split & Model & MSE & MAE & L2\\
\midrule
\multirow{5}{*}{Val.}
 & \piHalf\ baseline & 0.256338 & 0.138628 & 1.153725\\
 & \method & \best{0.249276} & \best{0.134522} & \best{1.128938}\\
 & GR00T N1.6 & 0.278784 & 0.162035 & 1.285486\\
 & StarVLA & 0.258356 & 0.163372 & 1.262874\\
 & SmolVLA & 0.257650 & 0.147754 & 1.182720\\
\midrule
\multirow{5}{*}{Test}
 & \piHalf\ baseline & 0.248680 & 0.134119 & 1.104541\\
 & \method & 0.244223 & \best{0.130593} & \best{1.071214}\\
 & GR00T N1.6 & 0.251833 & 0.152173 & 1.183549\\
 & StarVLA & \best{0.237490} & 0.149984 & 1.157864\\
 & SmolVLA & 0.252837 & 0.141363 & 1.133300\\
\bottomrule
\end{tabular}

\vspace{2pt}
\setlength{\tabcolsep}{2pt}
\begin{tabular}{@{}lcc@{}}
\toprule
Metric & Validation gain [95\% CI] & Test gain [95\% CI]\\
\midrule
MSE & +2.76 [$-1.92$, +7.23] & +1.79 [$-1.60$, +5.33]\\
MAE & +2.96 [+0.74, +5.15] & +2.63 [+0.82, +4.48]\\
L2 & +2.15 [$-0.65$, +4.89] & +3.02 [+0.94, +5.07]\\
\bottomrule
\end{tabular}
\end{table}

\subsection{Cross-Family References}

The GR00T reference has higher MAE and L2 than both \piHalf\ models on both
splits.  Relative to the \piHalf\ baseline, its
validation and test MAE are 16.89\% and 13.46\% higher; paired-window bootstrap
intervals for the increase are [12.50, 21.62]\% and [9.97, 17.14]\%.
StarVLA's test MAE and L2 are 11.83\% and 4.83\% higher than the \piHalf\
baseline.  SmolVLA obtains validation/test MAE of 0.147754/0.141363 and L2 error of
1.182720/1.133300.  \method\ achieves the lowest MAE and L2 on both splits
across all five policies, demonstrating that its predictive temporal structure
provides a strong advantage over multiple VLA families rather than only over
its direct \piHalf\ baseline.

\subsection{Action-Group Analysis}

The strongest group-level effect is end-effector roll/pitch/yaw (RPY):
validation MAE improves by 4.12\% and test MAE by 3.78\%
(Table~\ref{tab:groups}).  End-effector position improves by 1.03\% on test,
and arm error also improves on both splits.  The concentration of gains in
end-effector orientation shows that future/event supervision is particularly
effective at preparing the gripper pose for moving-object contact.

\begin{table}[t]
\caption{Pooled action-group MAE.  Parentheses show relative improvement.}
\label{tab:groups}
\centering
\small
\begin{tabular}{@{}llrr@{}}
\toprule
Split & Group & Baseline & \method\\
\midrule
\multirow{5}{*}{Val}
 & base XY & 0.016555 & 0.016176 (+2.30\%)\\
 & base yaw & 0.008367 & 0.008367 (+0.00\%)\\
 & EE position & 0.023643 & 0.023479 (+0.69\%)\\
 & EE RPY & 0.478830 & \best{0.459112 (+4.12\%)}\\
 & arm & 0.095593 & 0.094838 (+0.79\%)\\
\midrule
\multirow{5}{*}{Test}
 & base XY & 0.015229 & 0.015642 ($-2.71$\%)\\
 & base yaw & 0.003279 & 0.003279 ($-0.01$\%)\\
 & EE position & 0.023745 & 0.023499 (+1.03\%)\\
 & EE RPY & 0.475728 & \best{0.457734 (+3.78\%)}\\
 & arm & 0.087678 & 0.087318 (+0.41\%)\\
\bottomrule
\end{tabular}
\end{table}

\subsection{Inference-Time Component Ablation}

We additionally perform a controlled deployment-path intervention on the
\method\ checkpoint (Table~\ref{tab:ablation}).  Each row uses the
same checkpoint weights; the indicated condition is zeroed or injection path
is removed only at inference.  We evaluate 96 matched test windows and a
critical subset of 48 matched grasp/release-transition windows (phase 1), with
identical initial action noise across rows.  This smaller study isolates whether
the trained policy uses each condition.

\begin{table}[H]
\caption{Offline inference-time component ablation.  Overall metrics use 96
matched test windows; transition metrics use 48 matched phase-1
windows.  Lower is better.  All \method\ variants share weights and differ only
by component masking at inference.}
\label{tab:ablation}
\centering
\scriptsize
\setlength{\tabcolsep}{2pt}
\begin{tabular}{@{}lrrr@{}}
\toprule
Method & \shortstack{Overall\\MAE $\downarrow$} &
\shortstack{Transition\\MAE $\downarrow$} &
\shortstack{Transition\\EE-RPY MAE $\downarrow$}\\
\midrule
\piHalf\ baseline & 0.122124 & 0.150140 & 0.501144\\
\method\ w/o future condition & 0.128787 & 0.159982 & 0.498349\\
\method\ w/o phase/time condition & 0.134846 & 0.160672 & 0.497727\\
\method\ slow path only & 0.136006 & 0.156587 & 0.494265\\
\method\ fast path only & 0.127140 & 0.161749 & \best{0.479751}\\
\method & \best{0.118354} & \best{0.145320} & 0.482656\\
\bottomrule
\end{tabular}
\end{table}

The full model has the lowest overall and transition MAE.  Removing the future
condition increases transition MAE by 10.1\%, while removing phase/time
conditioning increases it by 10.6\%.  Restricting conditioning to either path
also degrades transition MAE, supporting complementary slow- and fast-path use.
Together, these interventions show that predictive future features, explicit
event timing, and dual-path conditioning are all actively used by the trained
policy, validating the central architectural choices of \method.

\subsection{Deployment Efficiency}

Pooled validation latency increases from 66.69 to 68.65 ms (+2.93\%), and test
latency from 66.97 to 68.62 ms (+2.46\%).  This small overhead is consistent
with the 1.251M added parameters and five extra slow-prefix tokens.  GR00T takes
133.86 and 132.12 ms on validation and test, respectively, while StarVLA takes
101.72 and 97.24 ms; both use four denoising steps.  Thus, \method\ combines the
strongest reported action accuracy and real-robot performance with substantially
lower latency than the cross-family references.  This efficiency follows
directly from the teacher--student design: the video-scale WAM supplies
predictive supervision offline, while deployment requires only a lightweight
1.251M-parameter temporal module and five additional prefix tokens.

\subsection{Quantitative Real-Robot Evaluation}

Table~\ref{tab:real_robot_tasks} reports the first closed-loop campaign.
\method\ attains the highest grasp success in both tasks.  On the stationary
task, it succeeds in 73/90 trials (81.1\%), 11 successes and 12.2 percentage
points above the next-best StarVLA result.  On the slow-speed moving task, it
succeeds in 53/90 trials (58.9\%), 20 successes and 22.2 points above the
next-best \piHalf\ baseline.  Relative to \piHalf, the absolute advantage grows
from 15.6 points when the target is stationary to 22.2 points when target motion
introduces an interception horizon.  This widening gap is consistent with the
intended role of the distilled future code: its benefit is largest when a
current-observation policy must extrapolate where and when contact will occur.

\begin{table}[H]
\caption{Real-robot outcomes for stationary and slow-speed moving grasp tasks.
All entries are percentages; higher success and lower failure are better.}
\label{tab:real_robot_tasks}
\centering
\scriptsize
\begin{tabular}{@{}lr@{}}
\toprule
\multicolumn{2}{c}{\textbf{Stationary grasp task}}\\
\cmidrule(lr){1-2}
Method & Grasp success\\
\midrule
\piHalf & 65.56\%\\
GR00T N1.6 & 46.67\%\\
StarVLA & 68.89\%\\
SmolVLA & 63.33\%\\
\method & \best{81.11\%}\\
\bottomrule
\end{tabular}

\vspace{4pt}

\begin{tabular}{@{}lrrrr@{}}
\toprule
\multicolumn{5}{c}{\textbf{Moving grasp task}}\\
\cmidrule(lr){1-5}
Method & Success & \shortstack{Early\\failure} &
\shortstack{Late\\failure} & \shortstack{Contact-pose\\error}\\
\midrule
\piHalf & 36.67\% & 21.11\% & 26.67\% & 15.56\%\\
GR00T N1.6 & 12.22\% & 27.78\% & 35.56\% & 24.44\%\\
StarVLA & 28.89\% & 32.22\% & 18.89\% & 20.00\%\\
SmolVLA & 32.22\% & 25.56\% & 27.78\% & 14.44\%\\
\method & \best{58.89\%} & \best{17.78\%} & \best{16.67\%} & \best{6.67\%}\\
\bottomrule
\end{tabular}
\end{table}

\method\ is the only policy with the best rate in all three moving-task failure
categories. Relative to \piHalf, early, late, and contact-pose failures decrease
by 15.8\%, 37.5\%, and 57.1\%, respectively. The larger reductions in late and
contact-pose errors are especially diagnostic: they occur after approach has
begun, when success depends on predicting contact time and preparing gripper
orientation. Together with the strongest offline gain in end-effector RPY, this
localizes the benefit to contact timing and geometry rather than generic action
smoothing.

\subsection{Robustness to Belt Speed}

\begin{figure}[H]
\centering
\includegraphics[width=0.88\columnwidth]{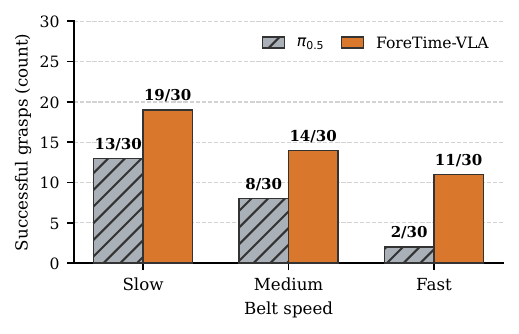}
\caption{\textbf{Real-robot grasp completions across belt speeds.} Successful
grasps out of 30 trials; labels give completed/total trials.}
\label{fig:real_robot_speed}
\end{figure}

The speed stress test in Fig.~\ref{fig:real_robot_speed} strengthens the
dynamic-manipulation result.  \method\ succeeds in 19 versus 13 trials at slow
speed, 14 versus 8 at medium speed, and 11 versus 2 at fast speed.  Across all
three settings, it completes 44/90 grasps compared with 23/90 for \piHalf,
adding 21 successful executions.  At fast speed, it completes more than five
times as many grasps as the baseline.

From slow to fast speed, \method\ drops from 19 to 11 successes, whereas
\piHalf\ drops from 13 to 2. Thus the absolute margin widens from six to nine
successful trials as the interception window contracts. The deployed policy
never observes future frames or runs the WAM, so this robustness must be carried
by the distilled causal history and event predictions.

\subsection{Temporal-Task Contribution}

An interception error changes both \emph{when} to close and \emph{how} to orient
the end effector. \method\ represents this coupling through a future code,
phase, and time-to-transition predicted from causal history, while all privileged
video remains offline. Its advantage over \piHalf\ grows from 15.6 percentage
points for stationary grasping to 22.2 points for slow motion and widens again
at fast speed. Together with the component ablations and orientation gains,
this trend supports predictive temporal conditioning, rather than generic
action smoothing, as the source of improvement.

\subsection{Qualitative Real-Robot Rollout}

\begin{figure}[H]
\centering
\includegraphics[width=\columnwidth]{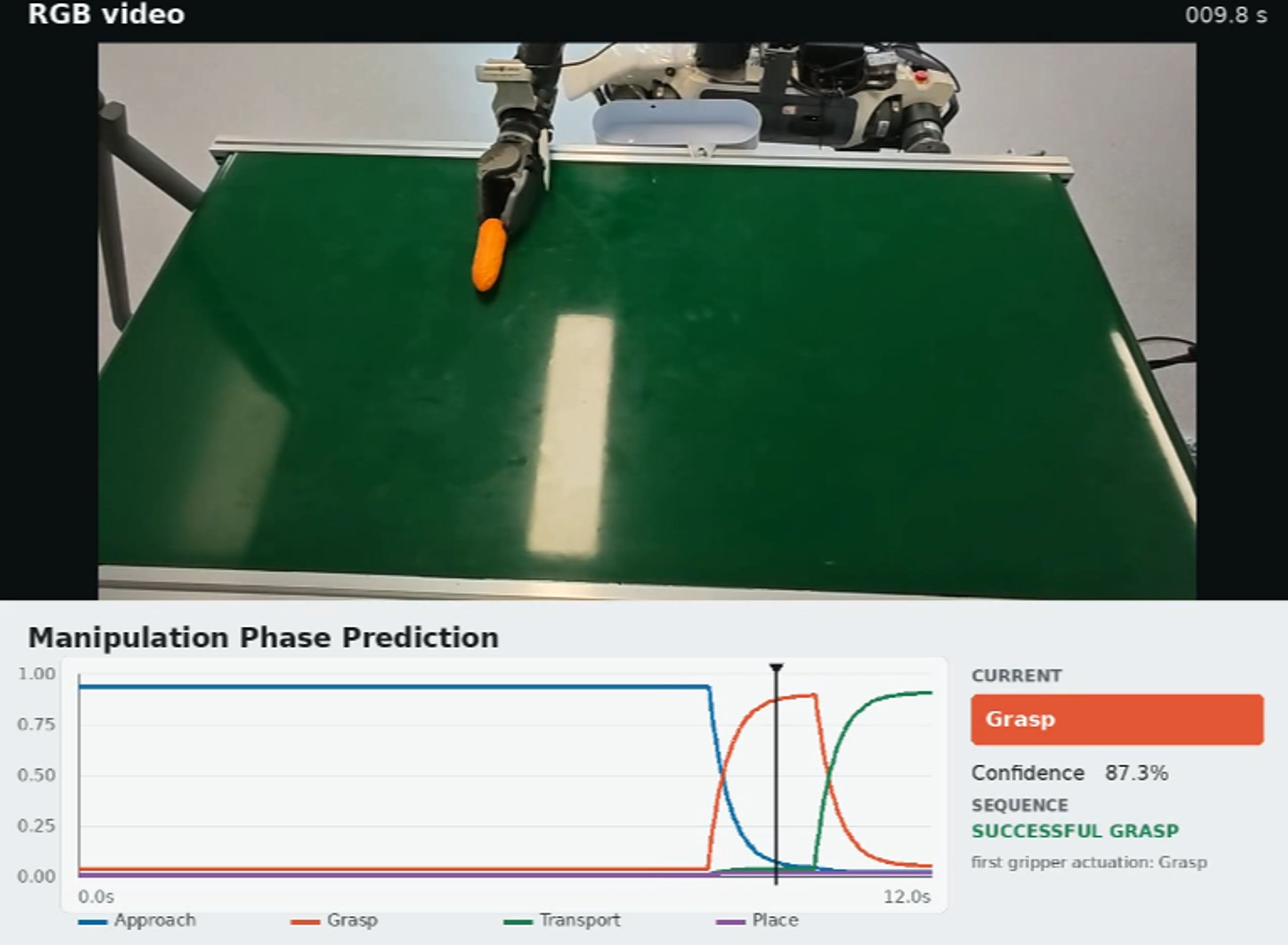}
\caption{\textbf{Qualitative real-robot grasp and phase prediction.}
\emph{Top:} grasp of the moving target. \emph{Bottom:} causal phase
probabilities; at $9.8$~s, grasp has 87.3\% confidence before transport rises
after acquisition.}
\label{fig:real_robot_grasp}
\end{figure}

Figure~\ref{fig:real_robot_grasp} shows one successful rollout. The phase trace
remains in approach while tracking the moving object. At $t=9.8$~s, the gripper
closes and the phase head assigns 87.3\% probability to grasp; the subsequent
rise in transport probability yields a coherent approach--grasp--transport
sequence rather than a static scene cue.

\subsection{Evidence Synthesis}

The evidence supports a consistent mechanism.  \method\ lowers MAE/L2 on both
splits; masking the future code, phase/time signal, or either conditioning path
worsens transition error; and its robot advantage grows as belt speed shortens
the interception horizon.  Together, these results connect the gain to temporal
conditioning rather than arbitrary capacity.  The effect is obtained with less
than 3\% latency overhead and no video generation at inference.  Our claim is
therefore narrower than domination of every scalar metric: test MSE favors
StarVLA, whereas stable MAE/L2 accuracy, causal deployment, and contact-relevant
robustness define the practical advantage of \method.

\section{Conclusion}

We presented \method, which transfers future-conditioned WAM structure into a
causal flow-matching VLA through an action-equivalent code, phase, and transition
horizon. It reduces test MAE/L2 by 2.63\%/3.02\% with under 3\% latency overhead,
reaches 81.1\% stationary and 58.9\% slow-moving grasp success, and completes
44/90 speed-stress trials versus 23/90 for \piHalf. The 11/30 versus 2/30 fast
speed result shows that compact future-token distillation is most useful when
reaction time is limited, without deploying the video teacher. Evidence remains
limited to one conveyor-belt embodiment and a fixed offline video backbone;
longer-horizon tasks, changing viewpoints, and broader action spaces are needed
to test generality. Overall, separating predictive supervision from test-time
generation provides the central practical benefit of the design.

\clearpage
\IEEEtriggeratref{20}
\bibliographystyle{IEEEtran}
\bibliography{references}

\end{document}